\documentclass[11pt]{article}

\usepackage[final]{acl}
\usepackage{amsmath}
\usepackage{amssymb}
\usepackage{graphicx}
\usepackage{overpic}
\usepackage{caption}
\usepackage{subcaption}
\usepackage{booktabs}
\usepackage[table]{xcolor}
\usepackage{float}
\usepackage{hyperref}
\definecolor{lightgrayrow}{gray}{0.93}
\usepackage{xcolor}
\newcommand{\cmark}{\textcolor{green!70!black}{\checkmark}}
\newcommand{\xmark}{\textcolor{red!70!black}{$\times$}}
\definecolor{lightgreen}{RGB}{232,245,233}

\usepackage{times}
\usepackage{latexsym}
\usepackage{enumitem}
\usepackage{multirow}
\setlist[itemize]{leftmargin=*}
\setlist[enumerate]{leftmargin=*}
\usepackage[T1]{fontenc}

\usepackage[utf8]{inputenc}

\usepackage{microtype}

\usepackage{inconsolata}

\usepackage{graphicx}
\setlist[itemize,enumerate,description]{nosep, itemsep=0.5mm}

\title{SciFlow-Bench: Evaluating Structure-Aware Scientific Diagram Generation via Inverse Parsing}

\author{
    Tong Zhang\textsuperscript{1}\thanks{Equal contribution.},
    Honglin Lin\textsuperscript{2}\footnotemark[1],
    Zhou Liu\textsuperscript{1}\thanks{Project Leader.},
    Chong Chen\textsuperscript{3},
    {\bf Wentao Zhang\textsuperscript{1,4,5}}\thanks{Corresponding author.} \\
    \textsuperscript{1}Peking University \quad
    \textsuperscript{2}Shanghai Jiao Tong University \\
    \textsuperscript{3}Huawei Cloud BU \quad
    \textsuperscript{4}Zhongguancun Academy \\
    \textsuperscript{5}Beijing Key Laboratory of Data Intelligence and Security (Peking University) \\
    \texttt{\{zhangtong25, zhouliu25\}@stu.pku.edu.cn, linhonglin@sjtu.edu.cn} \\
    \texttt{chenchong55@huawei.com, wentao.zhang@pku.edu.cn}
}

\begin{document}
\maketitle
\begin{abstract}
Scientific diagrams convey explicit structural information, yet modern text-to-image models often produce visually plausible but structurally incorrect results. Existing benchmarks either rely on image-centric or subjective metrics insensitive to structure, or evaluate intermediate symbolic representations rather than final rendered images, leaving pixel-based diagram generation underexplored. We introduce SciFlow-Bench\footnote{\url{https://github.com/Tong-0302/SciFlow-Bench}}, a structure-first benchmark for evaluating scientific diagram generation directly from pixel-level outputs. Built from real scientific PDFs, SciFlow-Bench pairs each source framework figure with a canonical ground-truth graph and evaluates models as black-box image generators under a closed-loop, round-trip protocol that inverse-parses generated diagram images back into structured graphs for comparison. This design enforces evaluation by structural recoverability rather than visual similarity alone, and is enabled by a hierarchical multi-agent system that coordinates planning, perception, and structural reasoning. Experiments show that preserving structural correctness remains a fundamental challenge, particularly for diagrams with complex topology, underscoring the need for structure-aware evaluation.
\end{abstract}

\section{Introduction}

\begin{figure}[t]
    \centering
    \includegraphics[width=\linewidth]{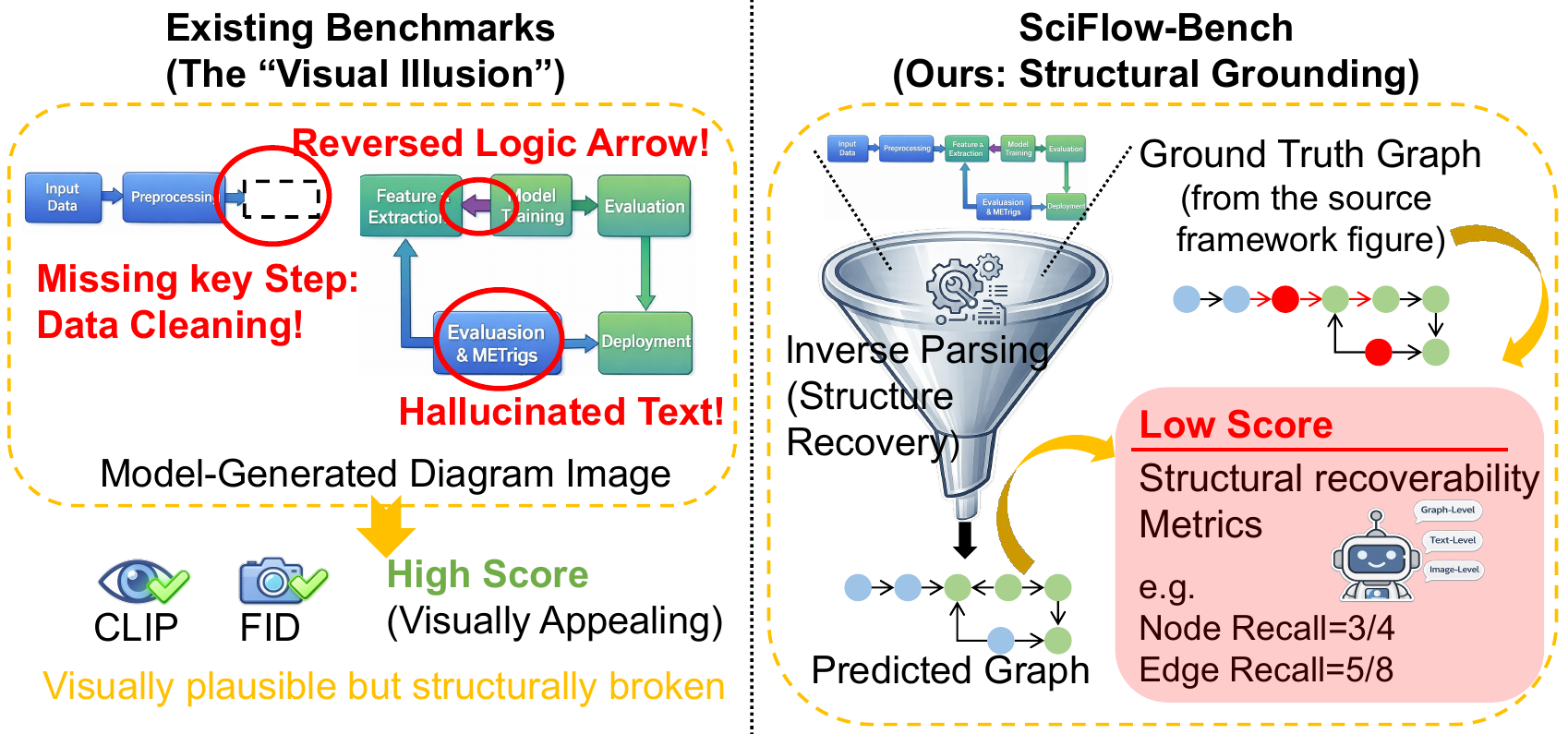}
    \caption{\textbf{Motivation and comparison between existing diagram benchmarks and SciFlow-Bench.} Left: Image-centric evaluation may assign high scores to visually plausible diagrams that contain structural errors. Right: SciFlow-Bench adopts a structure-first evaluation by inverse-parsing generated diagrams into graphs and measuring structural recoverability.}
    \label{fig:motivation}
  \vspace{-20pt}
\end{figure}

As structured artifacts, scientific diagrams encode methods, architectures, and workflows through explicit components and directed relations.
Unlike natural images, where aesthetic realism is often the primary objective, scientific diagrams are functional artifacts designed to convey explicit logical structure, including functional components, directed dependencies, and global information flow.
Faithful diagram generation therefore requires preserving underlying structure in addition to producing visually plausible layouts.
Despite rapid progress in text-to-image generation~\cite{rombach2022high, saharia2022photorealistic, betker2023improving}, modern diffusion-based models frequently struggle with scientific diagrams.
Given textual descriptions, they often produce images that appear visually coherent but contain structural errors, such as missing components, hallucinated labels, or semantic inconsistencies between textual annotations and arrow directions~\cite{cho2023visual, chen2023textdiffuser}.
These errors undermine the communicative function of diagrams and pose challenges for both automated analysis and human interpretation.

Most existing benchmarks fail to adequately capture such failures.
Many rely on image-level similarity metrics or subjective judgments that are largely insensitive to logical organization and directed structure~\cite{hessel2021clipscore, lin2024evaluating}.
Others evaluate intermediate symbolic representations rather than the final rendered diagram image~\cite{liang2025diagrameval, wang2025scisketch}, bypassing the ambiguities and failure modes introduced during pixel-level generation.
As a result, existing evaluation protocols lack a principled mechanism to test whether a generated diagram image can be structurally recovered into a coherent representation consistent with the source framework figure.
Recent benchmarks targeting scientific illustration, such as SridBench~\cite{chang2025sridbench}, further highlight the need for structure-aware evaluation under realistic scientific settings at scale.
As illustrated in Figure~\ref{fig:motivation}, this limitation gives rise to a visual illusion in current evaluation paradigms.
Diagrams that appear professional and well-organized at the pixel level may receive high scores under image-centric metrics, even when they contain critical logical errors such as reversed dependencies, missing methodological steps, or unsupported connections that break the underlying execution logic.
To address this gap, we introduce \textbf{SciFlow-Bench}, a new benchmark designed to evaluate scientific diagram generation under a structure-first criterion, directly from pixel-level outputs.

SciFlow-Bench is constructed from real-world scientific PDFs and treats models as black-box image generators evaluated solely on their final rendered diagram images.
Each source framework figure is paired with a canonical ground-truth graph encoding functional components and directed relations, which is automatically constructed by a hierarchical multi-agent system inspired by recent agentic reasoning frameworks~\cite{wu2024autogen, li2023camel}.
During evaluation, the same system inverse-parses generated diagram images into predicted graphs, forming a unified round-trip protocol from textual descriptions to pixels and back to structured representations in a deterministic pipeline.
This design evaluates structural recoverability by measuring whether a rendered diagram preserves the logical relations required for reliable automated or human reconstruction, rather than relying on visual similarity alone as the sole criterion.

Our contributions are summarized as follows:
\begin{itemize}
  \item We identify structural recoverability as an underexplored failure mode in pixel-based scientific diagram generation, exposing a visual illusion where visually plausible outputs fail to preserve recoverable graph-level structure, motivating a structure-first evaluation paradigm beyond image-centric similarity.
  \item We introduce SciFlow-Bench, a benchmark of 500 real-world scientific diagrams spanning five major research domains. SciFlow-Bench pairs source framework figures with canonical ground-truth graphs and operationalizes evaluation via a hierarchical multi-agent system (HMAS) that enables consistent round-trip graph construction and deterministic inverse parsing across evaluation stages.
  \item Through systematic and extensive evaluation of code-driven, diffusion-based, and autoregressive vision--language models, we empirically demonstrate a pronounced decoupling between visual fidelity and structural reasoning, particularly for diagrams with complex topology, underscoring structure-awareness as a central challenge for scientific multimodal generation in realistic scenarios.
\end{itemize}

\section{Related Work}

\textbf{Text-to-Image Models for Diagrams.}
Modern text-to-image generation is dominated by diffusion-based models, including latent diffusion~\citep{rombach2022high} and variants such as SDXL~\citep{podell2023sdxl}, PixArt-$\Sigma$~\citep{chen2024pixart}, and Qwen-Image~\citep{wu2025qwen}. Proprietary systems such as Gemini further improve prompt adherence through large language encoders~\citep{team2023gemini}. Despite strong visual fidelity, prior studies report limitations in modeling long-range dependencies, precise spatial relations, and text-object binding in scientific diagrams, often causing missing components or incorrect connections~\citep{lian2023llm,feng2024ranni}. These findings suggest that visual realism alone is insufficient for faithful diagram generation, and preserving structural relations remains a core challenge.

\textbf{Evaluation of Generated Images.}
Evaluation of image generation commonly relies on distribution-level metrics such as FID~\citep{heusel2017gans} and IS~\citep{salimans2016improved}, or image-text alignment measures such as CLIPScore~\citep{hessel2021clipscore}. While effective for assessing realism and coarse semantic relevance, these metrics remain largely insensitive to relational and topological structure.
Compositional benchmarks such as T2I-CompBench~\citep{huang2023t2i} partially address attribute binding and spatial relations. Domain-specific benchmarks for scientific images, such as SciGenBench~\cite{lin2026scientific}, further evaluate logical correctness and information utility. However, existing benchmarks still do not explicitly model the directed and hierarchical dependencies fundamental to many scientific diagrams. As a result, they cannot fully assess whether a generated diagram preserves the structural logic needed for reliable interpretation or faithful reconstruction.

\begin{figure*}[t]
\vspace{-15pt}
  \centering
  \includegraphics[width=0.95\textwidth]{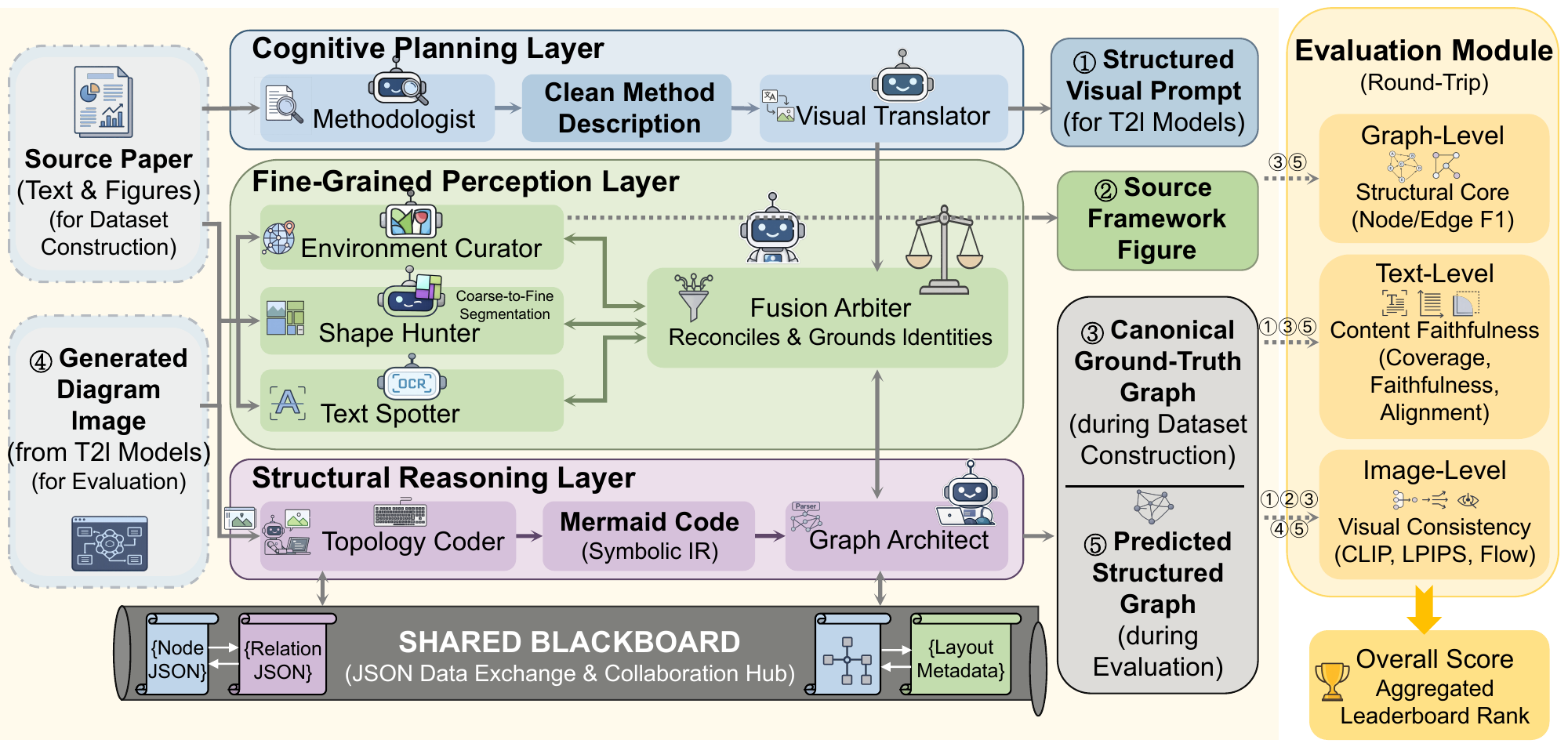}
  \caption{\textbf{Overview of the SciFlow-Bench framework.}
A unified round-trip evaluation pipeline based on a hierarchical multi-agent system that constructs canonical ground-truth graphs from source framework figures and recovers predicted graphs from generated diagram images.
By inverse-parsing pixel-level outputs into structured graphs, the same pipeline supports both dataset construction and structure-first evaluation via structural recoverability.}
  \vspace{-10pt}
  \label{fig:overview}
\end{figure*}

\textbf{Diagram Benchmarks.}
Prior diagram benchmarks typically emphasize a single modality, often sacrificing pixel-level assessment, explicit topology, or deterministic metrics. Benchmarks such as DocVQA~\citep{mathew2021docvqa}, ChartQA~\citep{masry2022chartqa}, and PlotQA~\citep{methani2020plotqa} focus on perception and reasoning over figures, rather than the structural fidelity of generated diagrams. Code-centric benchmarks such as DiagramEval~\citep{liang2025diagrameval} enable precise verification through symbolic layouts, but bypass pixel-level generation and its ambiguities. Thus, existing benchmarks do not jointly assess both visual and structural fidelity directly from pixels.

The closest prior work in evaluation philosophy is VPEval~\citep{cho2023visual}, which evaluates diagram relations through visual question answering. However, its reliance on manual annotations limits scalability and constrains coverage. In contrast, SciFlow-Bench evaluates pixel-based scientific diagram generation through inverse parsing, enabling scalable and deterministic comparison between predicted graphs reconstructed from generated diagram images and canonical ground-truth graphs derived from real source framework figures.

\section{SciFlow-Bench Framework}

\subsection{Problem Formulation and Overview}

SciFlow-Bench evaluates text-to-image models for scientific diagram generation under a structure-first paradigm, defining diagram quality by structural recoverability rather than visual similarity.
The task is formulated as a round-trip evaluation with two coupled stages.
Given a source paper $P$, a canonical ground-truth graph $G^{\ast}$ is constructed from its source framework figure using a graph-based representation adopted in prior work~\cite{johnson2015image, schuster2015generating}.
Given a method description from the same paper, a model generates a diagram image $I$, which is inverse-parsed into a predicted graph $\hat{G}$.
Performance is evaluated by comparing $\hat{G}$ and $G^{\ast}$ in structured graph space.
Figure~\ref{fig:overview} illustrates this round-trip paradigm.
To support canonical graph construction and inverse parsing, SciFlow-Bench adopts a three-layer hierarchical design consisting of cognitive planning, fine-grained perception, and structural reasoning.
This design follows a planning, perception, and reasoning decomposition adopted in agentic frameworks~\cite{yao2022react, wang2023plan}.

\subsection{Hierarchical Multi-Agent Pipeline}

The hierarchical multi-agent pipeline serves as the evaluation infrastructure of SciFlow-Bench, providing a unified and deterministic mechanism for both canonical graph construction and inverse parsing of generated diagram images.
The same pipeline is applied consistently during dataset construction and benchmark evaluation, ensuring strict consistency and reproducibility across all experimental stages throughout the framework.

\textbf{Cognitive Planning.}
The cognitive planning layer consists of two agents.
The \emph{Methodologist} extracts method-related passages from a source paper $P$, producing a normalized method description.
The \emph{Visual Translator} converts this description into a structured visual prompt specifying diagram components, relations, and stylistic constraints.
This prompt is shared across all text-to-image models without manual tuning, ensuring consistency.

\textbf{Fine-Grained Perception.}
The perception layer adopts a parallel multi-branch design followed by centralized fusion.
Three agents, the \emph{Environment Curator}, \emph{Shape Hunter}, and \emph{Text Spotter}, operate concurrently on either a source framework figure during dataset construction or a generated diagram image $I$ during evaluation.
The Environment Curator infers global layout flow.
The Shape Hunter performs hierarchical coarse-to-fine visual segmentation to recover functional regions and node-level components~\cite{lin2017feature}.
In parallel, the Text Spotter applies OCR to extract textual elements missed by visual segmentation.
All perceptual outputs are written to a shared blackboard following the classical blackboard architecture~\cite{penny1986blackboard}.
The \emph{Fusion Arbiter} integrates these signals by resolving overlaps, grounding identities via spatial alignment, and removing redundant elements, yielding a consolidated set of grounded nodes.

\textbf{Structural Reasoning.}
The structural reasoning layer converts grounded perceptual entities into explicit topology.
The \emph{Topology Coder} emits a symbolic intermediate representation in Mermaid that specifies directed relations under a strict ``what-you-see-is-what-you-write'' policy, enforcing explicit connectivity decisions and reducing unsupported relations~\cite{li2023evaluating, chen2022program}.
The \emph{Graph Architect} parses this representation and grounds abstract identifiers to concrete node instances, yielding a structured graph with nodes, edges, hierarchical grouping, and layout metadata.
The resulting graph is instantiated as the canonical ground-truth graph $G^{\ast}$ during dataset construction or as the predicted graph $\hat{G}$ during evaluation.

\begin{table*}[t]
\vspace{-15pt}
\centering
\small
\renewcommand{\arraystretch}{1.15}
\setlength{\tabcolsep}{6pt} 
\begin{tabular}{lcccccc}
\toprule
Domain & Node Prec. & Node Rec. & \textbf{Node F1} & Edge Prec. & Edge Rec. & \textbf{Edge F1} \\
\midrule
Computer Vision          & 0.88 & 0.93 & \textbf{0.89} & 0.65 & 0.67 & \textbf{0.65} \\
NLP                      & 0.92 & 0.97 & \textbf{0.94} & 0.77 & 0.86 & \textbf{0.81} \\
Machine Learning Theory  & 0.87 & 0.92 & \textbf{0.89} & 0.58 & 0.72 & \textbf{0.62} \\
Integrated Circuits      & 0.93 & 0.96 & \textbf{0.94} & 0.74 & 0.79 & \textbf{0.76} \\
Robotics                 & 0.83 & 0.96 & \textbf{0.88} & 0.69 & 0.81 & \textbf{0.72} \\
\rowcolor{lightgreen}
\textbf{Overall}         & \textbf{0.89} & \textbf{0.95} & \textbf{0.91} & \textbf{0.69} & \textbf{0.77} & \textbf{0.71} \\
\bottomrule
\end{tabular}
\caption{\textbf{Annotation reliability between HMAS and human verification.}
Agreement between the automated HMAS pipeline and human-verified ground-truth graphs on 100 source framework figures.
Node-level and edge-level Precision, Recall, and F1 are computed based on exact identity consistency to quantify annotation reliability.
These scores are reported separately from benchmark evaluation metrics.}
\label{tab:human_verification}
\end{table*}

\begin{table*}[t]
\vspace{-5pt}
\centering
\small
\renewcommand{\arraystretch}{1.15}
\setlength{\tabcolsep}{6pt}
\begin{tabular}{lcccccc}
\toprule
Image Distribution & Node Prec. & Node Rec. & \textbf{Node F1} & Edge Prec. & Edge Rec. & \textbf{Edge F1} \\
\midrule
Real Paper Images                 & 0.89 & 0.95 & \textbf{0.91} & 0.69 & 0.77 & \textbf{0.71} \\
Graphviz (GPT-4o$\rightarrow$DOT) & 0.95 & 0.98 & \textbf{0.96} & 0.86 & 0.92 & \textbf{0.89} \\
Qwen-Image                        & 0.86 & 0.88 & \textbf{0.87} & 0.67 & 0.71 & \textbf{0.68} \\
Gemini 3 Pro Image                & 0.88 & 0.92 & \textbf{0.90} & 0.65 & 0.75 & \textbf{0.70} \\
\bottomrule
\end{tabular}
\caption{\textbf{Generalization of HMAS across real and generated diagram distributions.}
Human-annotated reference graphs are compared with HMAS outputs under the same node-level and edge-level Precision, Recall, and F1 protocol used in Section~\ref{sec:annotation_quality}. The results show that HMAS remains stable across both real and generated distributions.}
  \vspace{-10pt}
\label{tab:hmas_generated_dist}
\end{table*}

\subsection{Structure-Aware Evaluation}
\label{sec:structure_eval}

Generated diagram images are evaluated by comparing reconstructed structured representations against canonical ground-truth graphs, rather than relying on pixel-level similarity alone.
All metrics operate in structured graph space, are normalized to the range $[0,1]$, and fall into three categories, namely graph-level, text-level, and image-level metrics, which capture structural correctness, semantic faithfulness, and overall visual consistency.

Graph-level evaluation assesses whether the topology recovered from a generated diagram image matches the canonical graph.
Since predicted and reference graphs do not share identifiers, nodes are matched based on semantic similarity of textual descriptions, with fallback to coarse type information when text is unavailable.
Node-level precision and recall measure component recovery, while edge-level evaluation focuses on directed dependencies.
A predicted edge is considered correct only if its endpoints correspond to a valid directed relation in the canonical graph, explicitly penalizing missing, reversed, or unsupported connections.
Structural correctness is summarized using node- and edge-level precision, recall, and F1 scores, with greater emphasis placed on relational accuracy.

Text-level and image-level evaluations capture complementary aspects beyond topology.
Text-level evaluation examines whether the predicted graph reflects the structured visual prompt by assessing component recovery, prompt support to penalize hallucinated elements, and higher-level alignment in module presence, hierarchical organization, and overall process flow.
Image-level evaluation focuses on visual properties not fully explained by topology alone, measuring semantic relevance to the prompt using CLIP-based similarity, perceptual and coarse layout similarity to the source framework figure using LPIPS, and visual flow consistency related to arrow directionality and spatial ordering using a vision-language model.

The final leaderboard score aggregates three evaluation metrics, with graph-level correctness receiving the highest emphasis.
This structure-first design prevents visually plausible but structurally inconsistent diagrams from achieving high scores.
Concrete metric definitions, weighting choices, and implementation details are provided in Appendix~\ref{sec:appendix_eval}.

\section{Dataset and Validation}

\subsection{Data Collection}

The benchmark is constructed from real-world scientific framework figures collected from arXiv papers published in 2025.
Each benchmark instance is anchored to a single source paper and consists of a source framework figure together with a canonical ground-truth graph.
This pair serves as the structural reference for all downstream evaluation tasks.
Canonical ground-truth graphs are automatically constructed from source framework figures by the hierarchical multi-agent system.
We treat graph construction as an integral part of data collection rather than a separate preprocessing stage.
No manual filtering, redrawing, or diagram simplification is applied.
In particular, ambiguous or implicitly defined relations present in the original figures are preserved.
This design retains the natural diversity and structural complexity of scientific diagrams and ensures that all evaluations are conducted under consistent and reproducible conditions.

\subsection{Annotation Quality and Reliability}
\label{sec:annotation_quality}

We conduct a human verification study on a stratified subset of 100 source framework figures, evenly distributed across five domains.
For each figure, trained master’s-level annotators in computer science review and refine the canonical ground-truth graph produced by the automated pipeline, yielding a human-verified graph under the same annotation schema and guidelines.
Verification follows a human-in-the-loop protocol commonly adopted in recent diagram benchmarks~\cite{chang2025sridbench, liang2025diagrameval}.
Importantly, annotators edit the automatically extracted graph rather than constructing a graph from scratch.
This design preserves node and edge identities and enables identity-consistent comparison.
Verification is performed using a custom web-based interface (Appendix~\ref{sec:appendix_interface}) for efficient and consistent graph editing.

Agreement is measured based on exact identity consistency.
Nodes and edges retained after verification are treated as correct, while removed or newly added elements are counted as mismatches.
Node-level and edge-level precision, recall, and F1 scores are reported following standard graph-structured evaluation practices~\cite{gao2010survey}.
As shown in Table~\ref{tab:human_verification}, the automated pipeline achieves strong agreement with human verification, with an overall Node-F1 of 0.91 and Edge-F1 of 0.71.
Node recall remains high across domains, indicating that most expert-confirmed components are already recovered automatically.
Edge-level performance is lower, reflecting the inherent difficulty of recovering directed relations in complex scientific diagrams.
Representative discrepancy cases are analyzed qualitatively in Appendix~\ref{sec:appendix_discrepancy}.

\begin{figure*}[t]
\vspace{-20pt}
  \centering
  \begin{minipage}{0.30\textwidth}
    \centering
    \includegraphics[height=3.6cm,keepaspectratio]{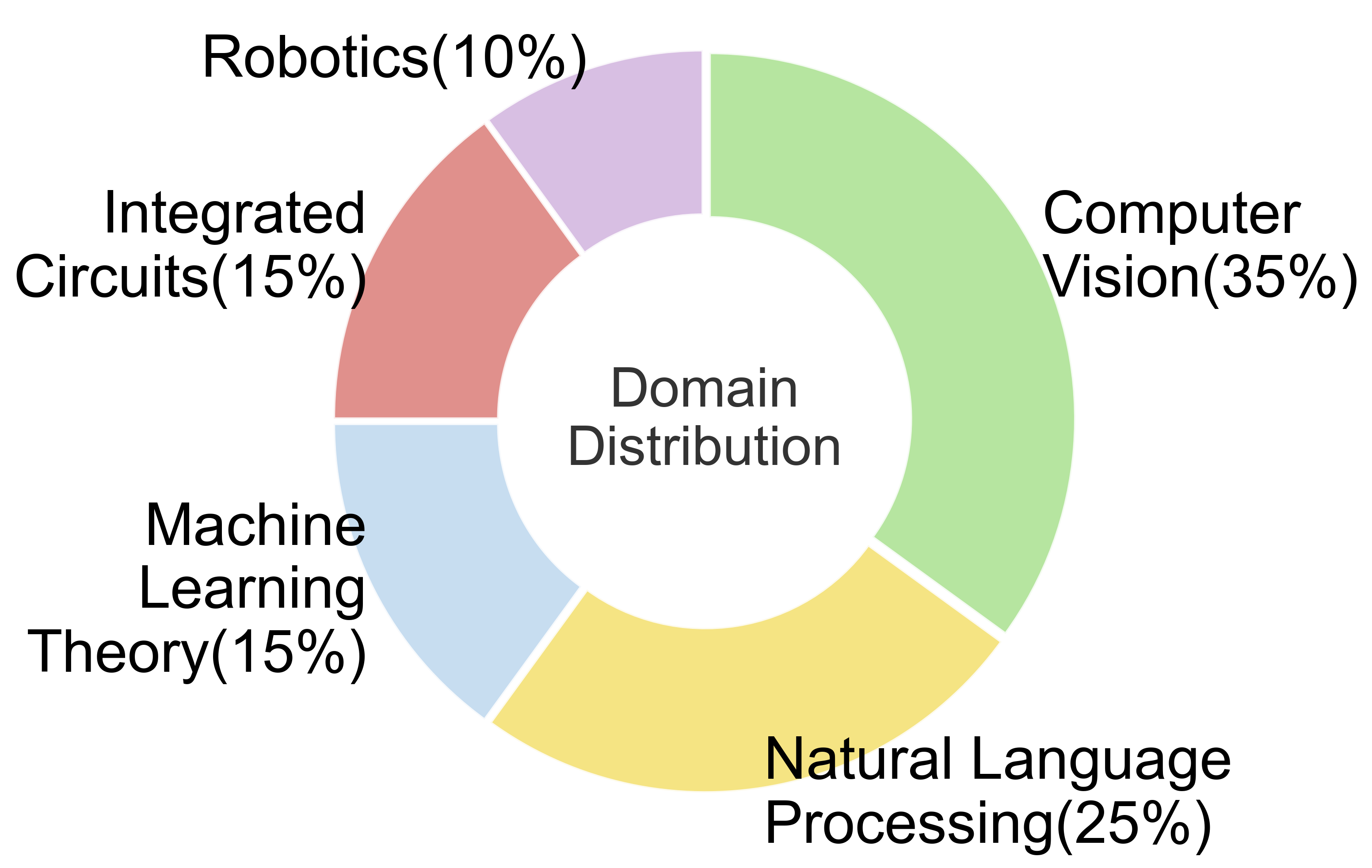}\par
    (a)
  \end{minipage}
  \hspace{6pt}
  \begin{minipage}{0.30\textwidth}
    \centering
    \includegraphics[height=3.6cm,keepaspectratio]{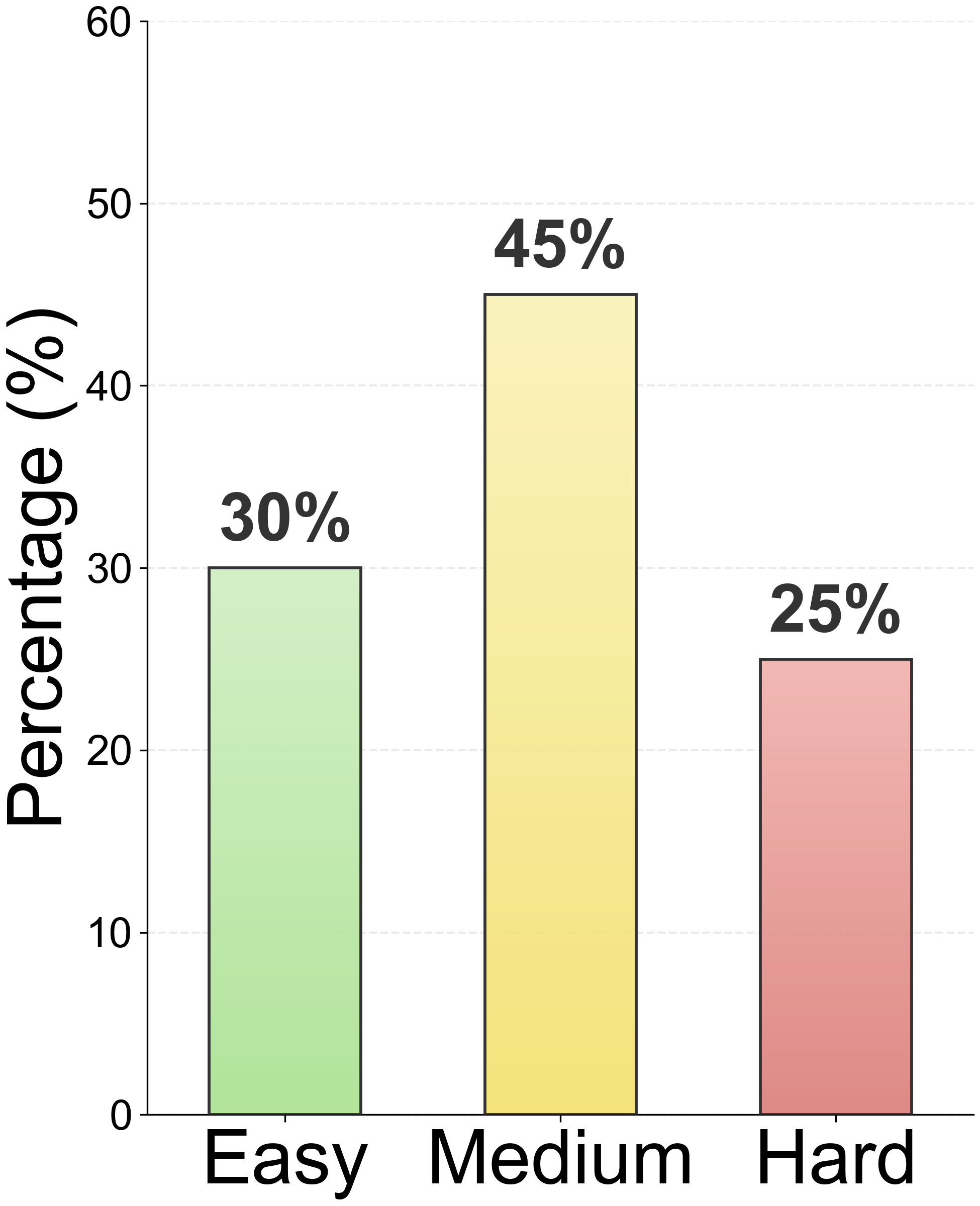}\par
    (b)
  \end{minipage}
  \hspace{6pt}
  \begin{minipage}{0.30\textwidth}
    \centering
    \includegraphics[height=3.6cm,keepaspectratio]{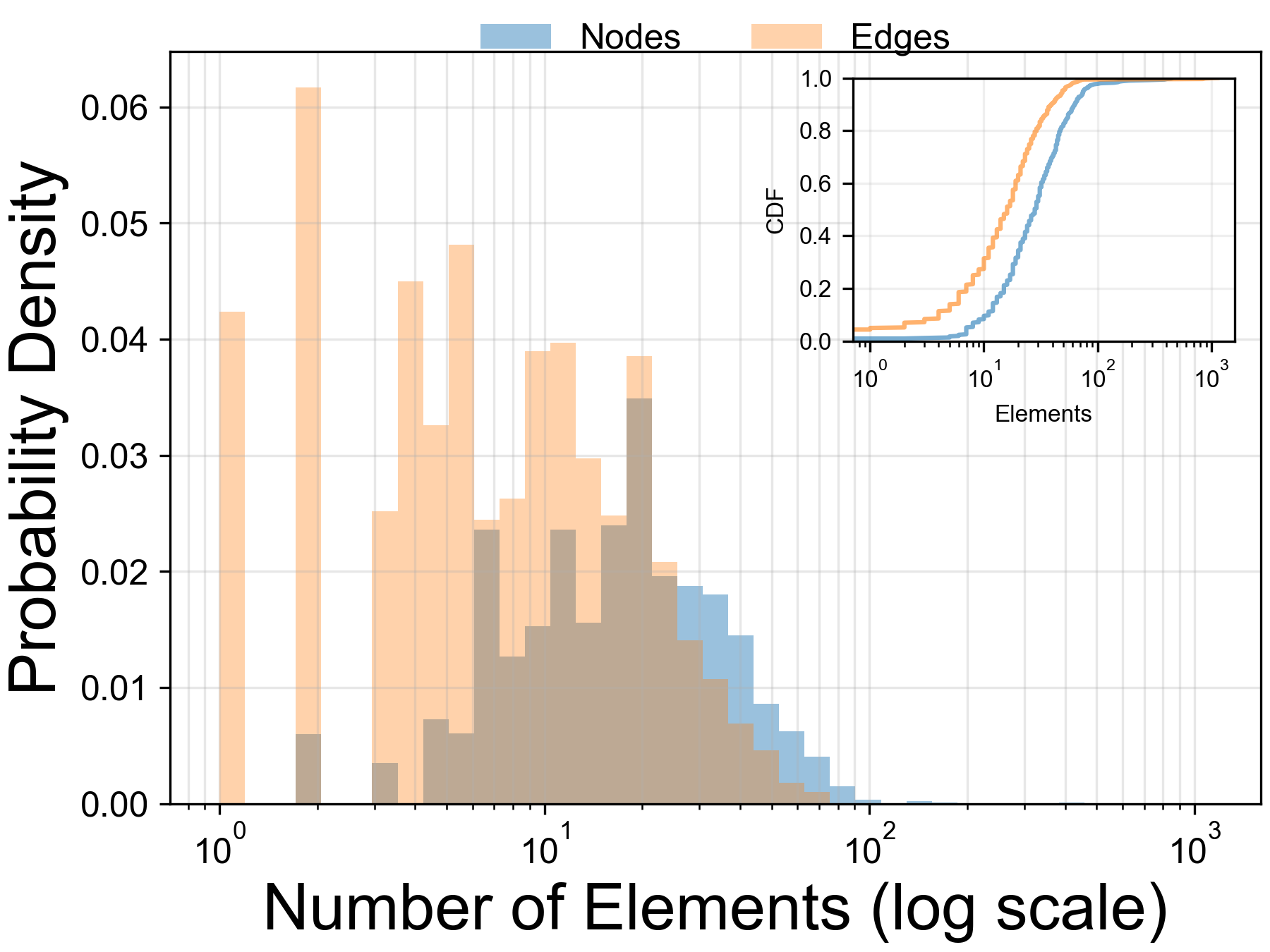}\par
    (c)
  \end{minipage}

  \caption{\textbf{Benchmark statistics of SciFlow-Bench.}
(a) Domain distribution across research areas.
(b) Distribution of structural difficulty levels defined by graph size and relational complexity.
(c) Long-tailed distributions of node and edge counts, reflecting substantial structural complexity in real-world scientific diagrams.}
  \vspace{-2pt}
  \label{fig:benchmark_stats}
\end{figure*}

\begin{table*}[t]
\centering
\scriptsize
\renewcommand{\arraystretch}{1.15}
\resizebox{\textwidth}{!}{
\begin{tabular}{lccccc}
\toprule
Benchmark & Evaluated Paradigm & Pixel-Level Eval & Explicit Topology & Multi-Modal Alignment & Deterministic Metrics \\
\midrule
MermaidBench & Code & \xmark & \cmark & \xmark & \cmark \\
DiagramEval & Code & \xmark & \cmark & \xmark & \cmark \\
SridBench & Diffusion & \cmark & \xmark & Model-based & \xmark \\
Paper2SysArch & Agent & \cmark & \xmark & \xmark & \xmark \\
\rowcolor{lightgreen}
SciFlow-Bench (Ours) & Diffusion & \cmark & \cmark & \cmark & \cmark \\
\bottomrule
\end{tabular}
}
\caption{\textbf{Comparison with recent benchmarks for scientific diagram generation and evaluation.}
SciFlow-Bench uniquely integrates pixel-level evaluation, explicit graph-level topology, and deterministic metrics within a single unified framework, enabling structure-first assessment directly from generated diagram images.}
  \vspace{-10pt}
\label{tab:benchmark_comparison}
\end{table*}

\subsection{HMAS on Generated Diagrams}

While Table~\ref{tab:human_verification} already validates HMAS on real source framework figures, benchmark evaluation is ultimately performed on model-generated diagrams. We therefore further examine whether HMAS remains stable under generated image distributions. Specifically, we sample 90 generated diagrams from three representative generated distributions, covering code-driven rendering, open-source image generation, and proprietary multimodal image generation, with 30 samples per source. Concretely, these distributions are instantiated using Graphviz (GPT-4o$\rightarrow$DOT), Qwen-Image, and Gemini 3 Pro Image, respectively. For each image, human annotators construct a reference graph under the same annotation schema used in Section~\ref{sec:annotation_quality}. We then compare HMAS outputs against these references using the same node-level and edge-level Precision, Recall, and F1 metrics.

As shown in Table~\ref{tab:hmas_generated_dist}, HMAS maintains stable behavior across both real and generated distributions. On structurally clean Graphviz diagrams, HMAS achieves Node-F1 of 0.96 and Edge-F1 of 0.89, providing an approximate upper bound when relations are explicit and visually unambiguous. On generated outputs from Qwen-Image and Gemini 3 Pro Image, edge-level agreement remains close to the real-paper baseline, with Edge-F1 scores of 0.68 and 0.70, respectively, compared with 0.71 on real source figures. These results suggest that the remaining edge-level errors primarily reflect ambiguity or structural collapse in the generated diagrams themselves, rather than systematic failure of the evaluator on generated inputs. This supplementary validation supports the use of HMAS as a stable and model-agnostic evaluator across both real and generated diagram distributions.

\subsection{Benchmark Statistics and Coverage}
\label{sec:benchmark_statistics}

Based on the verified dataset, we report overall statistics of SciFlow-Bench in Figure~\ref{fig:benchmark_stats}.
The benchmark contains a total of 500 instances constructed from source framework figures in arXiv papers published in 2025.
These instances span five major research areas, including Computer Vision (35\%), Natural Language Processing (25\%), Machine Learning Theory (15\%), Integrated Circuits (15\%), and Robotics (10\%), reflecting the distribution of contemporary AI research.
Structural difficulty is characterized by the size and relational complexity of the canonical graphs, including node count, branching density, and hierarchical organization.
Over 70\% of diagrams exhibit non-linear structures, indicating substantial topological complexity.
Node and edge count distributions further exhibit clear long-tailed behavior, consistent with prior observations of complex real-world systems~\cite{zipf2016human}.
Together, these statistics demonstrate that SciFlow-Bench captures both the diversity and structural richness of real scientific framework figures, providing a challenging testbed for evaluating structure-aware diagram generation.

\begin{table*}[t]
\vspace{-15pt}
\centering
\small
\setlength{\tabcolsep}{6pt}
\renewcommand{\arraystretch}{1.15}
\begin{tabular}{l l c c c c c c c}
\toprule
\textbf{Baseline Model} &
\textbf{Type} &
\multicolumn{4}{c}{\textbf{$S_{\text{graph}}$}} &
\textbf{$S_{\text{text}}$} &
\textbf{$S_{\text{image}}$} &
\textbf{$S_{\text{overall}}$} \\
\cmidrule(lr){3-6}
& &
\textbf{Easy} & \textbf{Medium} & \textbf{Hard} & \textbf{Avg.} &
\textbf{Avg.} & \textbf{Avg.} & \textbf{Avg.} \\
\midrule
Graphviz (GPT-4o$\rightarrow$DOT)
& Code-driven
& 0.076 & 0.098 & 0.098 & 0.091 & \textbf{0.359} & 0.463 & 0.283 \\
SDXL
& Diffusion
& 0.020 & 0.010 & 0.012 & 0.013 & 0.001 & 0.271 & 0.087 \\
PixArt-$\Sigma$
& Diffusion (DiT-based)
& 0.030 & 0.015 & 0.008 & 0.017 & 0.003 & 0.287 & 0.094 \\
Qwen-Image
& Diffusion (MMDiT)
& 0.059 & 0.090 & 0.090 & 0.081 & 0.243 & 0.411 & 0.229 \\
Gemini~2.5~Flash~Image
& Autoregressive VLM
& 0.077 & 0.119 & 0.117 & 0.106 & 0.315 & 0.458 & 0.274 \\
Gemini~3~Pro~Image
& Autoregressive VLM
& \textbf{0.079} & \textbf{0.129} & \textbf{0.135} & \textbf{0.116}
& 0.347 & \textbf{0.535} & \textbf{0.311} \\
\bottomrule
\end{tabular}
\caption{\textbf{Main results on SciFlow-Bench across structural difficulty levels.} Graph-level scores are reported for Easy, Medium, and Hard subsets defined in Section~\ref{sec:benchmark_statistics}, while text-level, image-level, and overall scores are averaged over all samples. Graphviz is a code-driven reference with deterministic prompt-to-code mapping.}
  \vspace{-10pt}
\label{tab:main_results}
\end{table*}

\subsection{Comparison with Prior Benchmarks}
\label{sec:benchmark_comparison}

Recent benchmarks proposed between 2024 and 2025 adopt diverse design choices in structural supervision and evaluation scope.
As summarized in Table~\ref{tab:benchmark_comparison}, code-centric benchmarks such as DiagramEval~\cite{liang2025diagrameval} and MermaidBench~\cite{wang2025scisketch} evaluate intermediate symbolic representations, enabling explicit topology supervision and deterministic metrics, but bypass pixel-level generation.
In contrast, image-space benchmarks such as SridBench~\cite{chang2025sridbench} operate directly on generated diagram images and emphasize visual or semantic alignment, yet do not explicitly recover or evaluate graph-level structure from pixels.
System-oriented approaches such as Paper2SysArch~\cite{guo2025paper2sysarch} focus on diagram construction pipelines rather than evaluation and do not provide benchmark-style, deterministic structural metrics.
SciFlow-Bench complements these efforts by unifying pixel-level evaluation, explicit graph-level topology, and deterministic structural metrics within a single round-trip protocol.
By evaluating structural recoverability directly from generated images, SciFlow-Bench occupies a previously unexplored position in the benchmark design space and enables aligned assessment across image, text, and topology.

\section{Experiments}

\subsection{Experimental Setup}

We evaluate representative baselines spanning code-driven layout generation and end-to-end pixel-based diagram synthesis under a unified evaluation protocol.
Across all experiments, the source paper, structured visual prompt, and inverse parsing pipeline are kept fixed, so performance differences arise solely from the generation models.
As a reference, we include a code-driven Graphviz pipeline, where GPT-4o converts the method description into DOT code and Graphviz deterministically renders the diagram image.
This baseline provides explicit topology and deterministic layout, serving as a diagnostic reference rather than a direct competitor to pixel-based generators, consistent with prior symbolic evaluation settings~\cite{liang2025diagrameval, wang2025scisketch}.
We further evaluate pixel-based models including Stable Diffusion XL~\cite{podell2023sdxl}, PixArt-$\Sigma$~\cite{chen2024pixart}, Qwen-Image~\cite{wu2025qwen}, and Gemini~2.5 Flash Image and Gemini~3 Pro Image~\cite{team2023gemini}.
All pixel-based models are treated as black-box generators and evaluated solely on their final rendered images under official default settings.
Structured visual prompts are produced automatically by the cognitive planning layer and shared across all models without manual tuning.
All generated images are inverse-parsed into predicted graphs using the same hierarchical multi-agent pipeline, ensuring fair and consistent comparison across generation paradigms.

\begin{table*}[t]
\vspace{-15pt}
\centering
\small
\renewcommand{\arraystretch}{1.15}
\setlength{\tabcolsep}{6pt}
\begin{tabular}{lccccccc}
\toprule
Configuration & Node P & Node R & Node-F1 & Edge P & Edge R & Edge-F1 & Diagnostic Score \\
\midrule
\rowcolor{lightgreen}
\textbf{Full Parsing Pipeline (Ours)} &
\textbf{0.90} & \textbf{0.96} & \textbf{0.92} &
\textbf{0.69} & \textbf{0.64} & \textbf{0.64} & \textbf{0.75} \\
w/o Shape Hunter & 0.81 & 0.79 & 0.78 & 0.38 & 0.34 & 0.34 & 0.52 \\
w/o Text Spotter & 0.85 & 0.48 & 0.59 & 0.32 & 0.24 & 0.24 & 0.38 \\
\bottomrule
\end{tabular}
\caption{\textbf{Ablation study of the parsing pipeline on the human-verified subset.} Pipeline outputs are compared to human-verified graphs using identity-consistent node- and edge-level Precision, Recall, and F1. The Diagnostic Score enables compact relative comparison among pipeline variants.}
  \vspace{-10pt}
\label{tab:ablation_hmas}
\end{table*}

\subsection{Main Results}
\label{sec:benchmark_results}

\subsubsection{Structural Performance Regimes}
\label{sec:model_regimes}

The evaluation results across diverse architectures reveal a clear hierarchy in structural reasoning performance.
As summarized in Table~\ref{tab:main_results}, evaluated systems cluster into three distinct performance regimes based on their overall scores, reflecting different capabilities in handling scientific topology.

\paragraph{Limits of vanilla diffusion generative models.}
Pure diffusion-based generators such as SDXL and PixArt-$\Sigma$ exhibit consistently weak structural recoverability in practice.
Across all difficulty levels, their graph-level scores remain close to zero, despite achieving moderate image-level relevance.
This pattern indicates that optimization for pixel-level distribution matching alone suffices to produce visually coherent layouts, but often fails to reliably preserve directed dependencies or stable node identities.
As a result, vanilla diffusion models remain ill-suited for scientific diagrams, where semantic correctness is primarily conveyed through explicit structure rather than appearance.

\paragraph{Emergent multimodal grounding.}In contrast, open-source models with extensive multimodal pre-training demonstrate markedly improved structural sensitivity.
Qwen-Image achieves a multi-fold improvement in average graph-level performance over PixArt-$\Sigma$ (from 0.017 to 0.081), substantially narrowing the gap to the code-driven Graphviz reference.
Notably, this improvement is consistent across Medium and Hard subsets, suggesting that large-scale vision--language alignment enables models to internalize implicit spatial and relational priors.
Although such models lack explicit symbolic reasoning steps, their behavior indicates a rudimentary form of structural grounding emerging from multimodal supervision alone.

\paragraph{Robustness of autoregressive architectures.}
Autoregressive vision--language models define the current upper bound across all evaluation dimensions in SciFlow-Bench settings.
Gemini~3~Pro~Image attains the highest overall score in Table~\ref{tab:main_results}, maintaining a substantial margin over the strongest open-source baselines.
Importantly, its graph-level performance improves monotonically with increasing structural difficulty, rising from 0.079 on Easy samples to 0.135 on the Hard subset.
This trend suggests that autoregressive scaling across modalities yields more robust internal representations of complex workflows, which are critical for synthesizing diagrams that remain structurally consistent under increasing topological complexity.

\subsubsection{Visual Structural Dissociation}
\label{sec:structural_limits}

Despite steady improvements in visual quality, SciFlow-Bench reveals a persistent gap between visual plausibility and structural recoverability.
This gap manifests consistently across model families, where image-level scores often dominate graph-level scores, indicating that visual realism alone remains a poor proxy for structural correctness.

Across all evaluated systems, visual relevance substantially outpaces recoverable structure.
For example, diffusion-based models such as SDXL produce visually plausible diagrams while exhibiting almost no recoverable topology under graph-level evaluation.
This discrepancy exposes a fundamental limitation of image-centric metrics such as CLIP or LPIPS: they reward visual resemblance without penalizing missing components, reversed dependencies, or unsupported relations that fundamentally alter scientific meaning in practice.
Analysis across structural difficulty levels further reveals a counterintuitive trend.
While diffusion-based models degrade steadily as complexity increases, advanced systems such as Gemini~3~Pro~Image exhibit improved structural recoverability on the Hard subset.
This behavior suggests that structurally complex diagrams are often accompanied by richer textual descriptions, which top-tier vision--language models can exploit to disambiguate relations more effectively.
Less capable models, by contrast, are overwhelmed by increased topological density and fail to maintain consistent connectivity. Appendix~\ref{sec:appendix_qualitative} further provides side-by-side qualitative comparisons across generation paradigms under identical structured prompts.

\subsubsection{Symbolic Generative Trade-offs}
\label{sec:symbolic_vs_generative}

Taken together, comparisons between symbolic and generative paradigms highlight a fundamental trade-off between logical determinism and visual expressiveness.
As a code-driven reference, Graphviz achieves the strongest text-level faithfulness due to its deterministic mapping from method descriptions to explicit topology.
However, this determinism constrains visual expressiveness and often results in rigid, less polished layouts in practice.
In contrast, pixel-based generative models produce diagrams that are visually refined and closer to publication quality.
Gemini~3~Pro~Image achieves the strongest image-level performance, but this flexibility comes with increased logical uncertainty.
Although its text-level alignment remains competitive, subtle structural inconsistencies, such as ambiguous edge attachments, are frequently introduced and visually smoothed over in the final rendering.

By grounding pixel-level outputs back into structured graph space, SciFlow-Bench provides a deterministic mechanism for exposing such uncertainty.
Together, these results motivate a structure-first evaluation axis that jointly considers visual realizability and topological correctness, preventing visually plausible but structurally invalid diagrams from achieving high rankings in practice.

\begin{table*}[t]
\vspace{-15pt}
\centering
\small
\renewcommand{\arraystretch}{1.15}
\setlength{\tabcolsep}{7pt}
\begin{tabular}{llcccc}
\toprule
\textbf{Baseline Model} & \textbf{Input Strategy} & \textbf{$S_{\text{graph}}$} & \textbf{$S_{\text{text}}$} & \textbf{$S_{\text{image}}$} & \textbf{$S_{\text{overall}}$} \\
\midrule
\rowcolor{lightgrayrow}
\multirow{2}{*}{Graphviz (GPT-4o$\rightarrow$DOT)} & SP & 0.091 & 0.359 & 0.463 & 0.283 \\
                                                    & NP & 0.048 & 0.182 & 0.221 & 0.140 \\[2pt]
\rowcolor{lightgrayrow}
\multirow{2}{*}{SDXL}                               & SP & 0.013 & 0.001 & 0.271 & 0.087 \\
                                                    & NP & 0.005 & 0.000 & 0.198 & 0.061 \\[2pt]
\rowcolor{lightgrayrow}
\multirow{2}{*}{PixArt-$\Sigma$}                    & SP & 0.017 & 0.003 & 0.287 & 0.094 \\
                                                    & NP & 0.007 & 0.001 & 0.215 & 0.068 \\[2pt]
\rowcolor{lightgrayrow}
\multirow{2}{*}{Qwen-Image}                         & SP & 0.081 & 0.243 & 0.411 & 0.229 \\
                                                    & NP & 0.038 & 0.115 & 0.284 & 0.135 \\[2pt]
\rowcolor{lightgrayrow}
\multirow{2}{*}{Gemini 2.5 Flash Image}             & SP & 0.106 & 0.315 & 0.458 & 0.274 \\
                                                    & NP & 0.046 & 0.155 & 0.312 & 0.159 \\[2pt]
\rowcolor{lightgrayrow}
\multirow{2}{*}{Gemini 3 Pro Image}                 & SP & 0.116 & 0.347 & 0.535 & 0.311 \\
                                                    & NP & 0.051 & 0.168 & 0.352 & 0.176 \\
\bottomrule
\end{tabular}
\caption{\textbf{Planning-layer ablation via prompt design.}
SP denotes the full Structured Prompt produced by the planning layer, while NP denotes a Naive Prompt that retains only the method description without explicit structural normalization. All models are evaluated under the same dataset, generation setting, and inverse-parsing protocol.}
\vspace{-10pt}
\label{tab:planning_ablation}
\end{table*}

\subsection{Ablation Study}

We conduct ablation studies to analyze the contribution of components in the hierarchical pipeline. Because layers serve different roles, we adopt two complementary strategies. For perception modules in the parsing backend, we perform direct removal on the human-verified subset to diagnose the parser. For the planning layer, which is task-defining rather than a drop-in module, we study its contribution through functional prompt ablation.

For the parsing pipeline, we evaluate three variants: without the Shape Hunter, without the Text Spotter, and the full pipeline. This study is conducted on the human-verified subset, and outputs are compared against human-verified graphs using identity-consistent node- and edge-level precision, recall, and F1 scores. Table~\ref{tab:ablation_hmas} shows that removing the Shape Hunter degrades structural completeness. Without hierarchical coarse-to-fine segmentation, the parser suffers from under-segmentation, leading to a sharp drop in edge recall and an overly sparse topology~\cite{lin2017feature}. Ablating the Text Spotter causes more severe degradation: node recall drops from $0.96$ to $0.48$, and the diagnostic score is nearly halved. Although some visual regions remain detectable, the lack of textual grounding prevents correct semantic identification and propagates to widespread edge-level failures~\cite{chen2023textdiffuser}. By contrast, the full pipeline achieves the most balanced performance across metrics. These results show that the Shape Hunter and Text Spotter provide complementary signals, and that their integration through the Fusion Arbiter is essential for constructing semantically grounded and topologically faithful graph representations.

We further study the planning layer through functional prompt ablation. Directly removing this layer would substantially change the input specification and make structural correctness under-defined. We therefore compare two input strategies: the full Structured Prompt (SP), which includes normalized structural intent and visual translation, and a Naive Prompt (NP), which retains only the method description and omits explicit structural normalization. All other factors are kept fixed. Table~\ref{tab:planning_ablation} shows that Structured Prompts consistently outperform Naive Prompts across all representative generators. The gain is especially clear in graph-level recovery, where Naive Prompts lead to lower $S_{\text{graph}}$ and reduced overall scores. This trend holds across instruction-following multimodal models, diffusion-based generators, and the code-driven Graphviz reference. These results indicate that the planning layer is more than a prompt engineering convenience: it explicitly specifies intended structure and enables a well-defined structure-first evaluation setting. Without it, the task becomes semantically under-specified, and model differences become less stable and less interpretable.

\section{Conclusion}

We present SciFlow-Bench, a structure-first benchmark for evaluating scientific diagram generation from pixel-level outputs. By pairing source framework figures with canonical ground-truth graphs and adopting a unified round-trip evaluation protocol, SciFlow-Bench enforces structural recoverability as the central criterion for diagram quality, rather than visual plausibility. Extensive experiments across layout-driven, open-source, and commercial text-to-image models show that preserving structural correctness remains a fundamental challenge, particularly for diagrams with complex topology. These results highlight a decoupling between visual fidelity and structural reasoning in multimodal generation models under realistic settings. By grounding evaluation in explicit graph-level structure reconstructed from generated images, SciFlow-Bench provides a principled and scalable framework for diagnosing structural failures that are invisible to image-centric metrics. We hope this benchmark helps establish structural recoverability as a core criterion for diagram generation and supports development of multimodal systems that can reason about scientific structure.

\section*{Limitations}

SciFlow-Bench adopts a structure-first evaluation paradigm and therefore does not explicitly assess fine-grained visual aesthetics such as stylistic preferences or rendering quality. In addition, the benchmark focuses on framework-style scientific diagrams with explicit components and directed dependencies, and does not cover all possible diagram types. We leave broader diagram coverage and the integration of aesthetic evaluation as future work.

\section*{Acknowledgments}

This work is supported by Fundamental and Interdisciplinary Disciplines Breakthrough Plan of the Ministry of Education of China (JYB2025XDXM113), National Natural Science Foundation of China (92470121, 62402016), National Key R\&D Program of China (2024YFA1014003), Zhongguancun Academy (C20250204, C20250602),  Beijing Major Science and Technology Project (Z251100008125043, Z251100008425023), and High-performance Computing Platform of Peking University.

\section*{Ethical Considerations}

SciFlow-Bench is constructed from publicly available scientific papers and framework figures and does not involve personal data or human subjects.
The benchmark is intended solely for research and evaluation purposes, focusing on structural properties of generated diagrams rather than the content or authorship of the source papers.
We do not foresee significant ethical risks arising from the use of this benchmark, and all results should be interpreted in a diagnostic context.

\bibliography{custom}

\clearpage
\appendix

\section{Evaluation Implementation Details}
\label{sec:appendix_eval}

This appendix provides detailed implementation details for the evaluation protocol described in Section~\ref{sec:structure_eval}.
All evaluation metrics strictly follow the definitions in the main text.
We describe concrete algorithmic choices, representation formats, aggregation strategies, and weighting schemes that ensure reproducibility and faithful assessment of structural recoverability from generated scientific diagram images.

\subsection{Graph-Level Metric Implementation}

Graph-level evaluation measures whether the logical topology reconstructed from a generated diagram image matches the canonical ground-truth graph derived from the source framework figure.
Because predicted and reference graphs do not share explicit identifiers, all comparisons are performed via semantic matching.

\paragraph{Node Representation and Matching.}
Each node is represented by a short textual description produced during inverse parsing, summarizing the semantic role of the corresponding visual or textual element.
Semantic similarity between node descriptions is computed using sentence-level embeddings.
Each description is encoded into a vector representation using a pretrained sentence embedding model, and similarity is measured using cosine similarity.
When textual descriptions are unavailable or unreliable, coarse node type labels are used as fallback representations to enable approximate semantic matching.

A reference node is considered recovered if it can be matched to at least one predicted node whose semantic similarity exceeds a fixed threshold.
This asymmetric matching criterion is robust to over-segmentation and duplicated predictions, which frequently occur in pixel-based diagram generation.

\paragraph{Edge Matching.}
Edges represent directed dependencies between nodes.
Edge evaluation follows a path-aware semantic matching protocol.
A predicted directed edge is considered correct if its source and target nodes can be semantically matched to a pair of nodes in the canonical graph such that a valid directed dependency exists between them.

To account for minor omissions of intermediate nodes in complex diagrams, reachability along the canonical graph topology is permitted.
This allows a predicted dependency to be considered correct if it preserves the overall directional logic, even when fine-grained intermediate steps are skipped.
Predicted edges with incorrect directionality, unsupported endpoints, or hallucinated relations are counted as errors.

\paragraph{Graph-Level Aggregation and Weights.}
Graph-level structural correctness is summarized using node-level and edge-level precision, recall, and F1 scores.
To reflect the central role of relational structure in scientific diagrams, edge-level F1 contributes a larger proportion to the graph-level score than node-level F1.

Specifically, node-level F1 contributes forty percent of the graph-level score, while edge-level F1 contributes sixty percent.
This weighting emphasizes preservation of directed dependencies over isolated component detection.
All graph-level aggregation weights are fixed a priori and shared across all evaluated models.

\subsection{Text-Level Metric Implementation}

Text-level evaluation measures consistency between the predicted graph and the structured visual prompt generated by the cognitive planning layer.
This evaluation assesses whether the generated diagram faithfully reflects the intended components and organization specified by the source paper.

\paragraph{Preprocessing and Filtering.}
To improve robustness to OCR noise and spurious detections, predicted node descriptions are filtered to exclude isolated characters, pure digits, and subfigure labels.
This preprocessing step reduces false positives that do not correspond to meaningful diagram components.

\paragraph{Coverage and Faithfulness.}
Semantic matching between prompt components and predicted graph nodes is performed using sentence embeddings and cosine similarity, following the same matching mechanism as node-level graph evaluation.
Coverage measures the fraction of prompt-specified components that are recovered in the predicted graph.
Faithfulness measures the fraction of predicted components that are supported by the structured visual prompt, penalizing hallucinated elements.

\paragraph{Alignment.}
Beyond component-wise matching, alignment captures higher-level structural consistency, including agreement in module presence, hierarchical organization, and overall process flow.
Alignment is assessed using a fixed vision-language model prompted to judge whether the predicted graph structure is globally consistent with the prompt specification.
The resulting score is normalized to the unit interval.
The vision-language model used for alignment evaluation is GPT-4o.
Importantly, this evaluator model is fixed across all experiments and is strictly decoupled from the generation models under evaluation.
In particular, GPT-4o is not used as a generation model in SciFlow-Bench and is distinct from proprietary generators such as Gemini~3~Pro~Image.
This design avoids self-evaluation and ensures consistent, model-agnostic judgment of structural alignment.

\paragraph{Text-Level Aggregation and Weights.}
The text-level score integrates coverage, faithfulness, and alignment.
Alignment contributes forty percent of the text-level score to emphasize global organizational consistency.
Coverage and faithfulness each contribute thirty percent, accounting for component-level recall and hallucination control.
All text-level aggregation weights are fixed and identical for all evaluated models.

\subsection{Image-Level Metric Implementation}

Image-level evaluation captures visual properties not fully explained by recovered topology or prompt alignment.
These metrics assess whether a generated diagram image is visually coherent, interpretable, and consistent with the intended structure at the pixel level.

\paragraph{Semantic Consistency.}
Semantic consistency between a generated diagram image and the structured visual prompt is measured using CLIP-based similarity.
The image and text are encoded into a shared embedding space using a fixed image encoder and text encoder, and similarity is computed via cosine similarity.
This score reflects high-level semantic alignment between visual content and the intended diagram description.

\paragraph{Perceptual and Layout Similarity.}
Perceptual similarity is measured using the Learned Perceptual Image Patch Similarity metric.
LPIPS compares deep feature representations extracted from multiple layers of a fixed convolutional network.
Lower LPIPS values indicate greater perceptual and coarse layout similarity.
For aggregation, this distance is inverted and normalized so that higher scores correspond to higher similarity.

\paragraph{Visual Flow Consistency.}
Visual flow consistency evaluates whether the generated diagram exhibits coherent directional and organizational structure.
This includes arrow directionality, spatial ordering of components, and grouping consistency.
A fixed vision-language model is prompted to assess whether the visual flow in the diagram supports a clear and interpretable process or dependency structure.
The resulting score is normalized to the unit interval.
The same fixed evaluator model, GPT-4o, is used for visual flow assessment across all models.
The evaluator operates under a deterministic prompt template and does not have access to model identities or generation metadata.
As the evaluator is independent of all evaluated generators, including Gemini~3~Pro~Image, this protocol mitigates potential evaluator bias while preserving consistent judgment criteria.

\paragraph{Image-Level Aggregation and Weights.}
The image-level score integrates semantic consistency, perceptual similarity, and visual flow consistency.
Semantic consistency contributes forty percent of the image-level score.
Visual flow consistency also contributes forty percent, reflecting its importance for interpretability and directional coherence.
Perceptual similarity contributes the remaining twenty percent as an auxiliary signal capturing coarse layout resemblance.
All image-level weights are fixed a priori and shared across all models.

\subsection{Overall Score Aggregation}

The final leaderboard score aggregates graph-level, text-level, and image-level scores.
Graph-level evaluation contributes forty percent of the overall score, reflecting the structure-first nature of the benchmark.
Text-level and image-level evaluations each contribute thirty percent, accounting for semantic faithfulness and visual consistency.
All aggregation weights are fixed across experiments and are not tuned per model or per domain.

\subsection{Reproducibility}

All evaluated models receive identical structured visual prompts without manual tuning.
All generated diagram images are rendered at a fixed resolution.
Canonical ground-truth graph construction and inverse parsing during evaluation both use the same hierarchical multi-agent parsing pipeline.
All thresholds, aggregation weights, models, prompts, and evaluation scripts are fixed across experiments to facilitate reproducibility and fair comparison.

\section{Human Verification Interface}
\label{sec:appendix_interface}

This appendix describes the web-based annotation interface used for human verification of canonical ground-truth graphs, as referenced in Section~\ref{sec:annotation_quality}.
The interface is designed to support efficient, identity-consistent refinement of automatically extracted graphs under a strict minimal-intervention protocol.
Its primary purpose is to enable reliable measurement of annotation quality while preserving comparability between automatic and human-verified graph representations.

\subsection{Design Objectives}

The interface is designed with three core objectives.

First, it enforces \emph{identity consistency}.
Human verification is performed by editing the output of the automated HMAS pipeline rather than constructing a graph from scratch.
This ensures that node and edge identifiers remain stable throughout the verification process, enabling exact identity-based comparison between automatic and human-verified graphs.

Second, it follows a \emph{minimal-intervention} principle.
Annotators are encouraged to make the smallest possible set of edits necessary to correct clear errors, prioritizing the removal of unsupported elements and the addition of missing but semantically necessary components.
This design prevents overfitting to individual annotator preferences and avoids introducing subjective reinterpretations of diagram structure.

Third, the interface is optimized for \emph{efficiency and consistency}.
By preloading all automatically extracted elements and providing constrained editing actions, the interface reduces annotation time while maintaining consistent verification behavior across annotators and domains.

\subsection{Interface Overview}

The interface presents annotators with a synchronized dual-view layout.
The source framework figure is displayed on one side, while the corresponding automatically extracted graph is shown on an editable graph canvas.
Nodes and directed edges produced by the HMAS pipeline are preloaded and visually selected by default.

Each node in the graph is associated with a unique identifier, textual description, and bounding box reference.
Edges are represented as directed connections between node identifiers.
This explicit linkage between visual elements and graph structure allows annotators to directly verify whether each extracted component and relation is supported by the source diagram.

A control panel lists all nodes and edges, enabling annotators to inspect elements individually, search by identifier or description, and perform batch operations.
This panel also records all edits performed during the verification process for auditing and analysis.

\subsection{Interaction Modes}

Annotators interact with the graph using two primary modes.

\paragraph{Select/Delete Mode.}
In this mode, annotators can select nodes or edges that are not supported by the source framework figure and mark them for exclusion.
Typical removal cases include spurious OCR artifacts, duplicated visual regions, or inferred relations that are visually ambiguous or unjustified.
Excluded elements are added to an internal exclusion list but remain logged for traceability.

\paragraph{Link Nodes Mode.}
This mode allows annotators to add missing directed relations by selecting a source node and a target node sequentially.
This operation is used sparingly, primarily to recover clear dependencies that are visually evident in the diagram but missed by automatic parsing.
Newly added edges are assigned fresh identifiers and explicitly marked as human-added elements.

Importantly, annotators are not permitted to freely redraw layouts, rename nodes arbitrarily, or restructure the graph wholesale.
These constraints enforce the minimal-intervention protocol and ensure that verification remains corrective rather than generative.

\subsection{Identity-Consistent Editing Protocol}

All nodes and edges extracted by the automated pipeline are included by default at the start of verification.
Annotators explicitly indicate unsupported elements by excluding them, rather than selectively including supported ones.
This default-inclusion strategy avoids confirmation bias and ensures that false positives are explicitly accounted for.

Elements retained after verification are treated as correct extractions.
Excluded elements correspond to false positives.
Newly added nodes or edges correspond to false negatives of the automated pipeline.
Because all edits operate directly on the original identifiers, node-level and edge-level precision and recall can be computed deterministically without post-hoc semantic matching or heuristic alignment.

This identity-consistent editing scheme allows annotation quality metrics reported in Table~\ref{tab:human_verification} to be interpreted as direct measures of the automated pipeline’s accuracy, rather than subjective agreement between independent annotations.

\begin{figure}[t]
  \centering

  \begin{minipage}[t]{0.95\columnwidth}
    \centering
    \includegraphics[width=\linewidth]{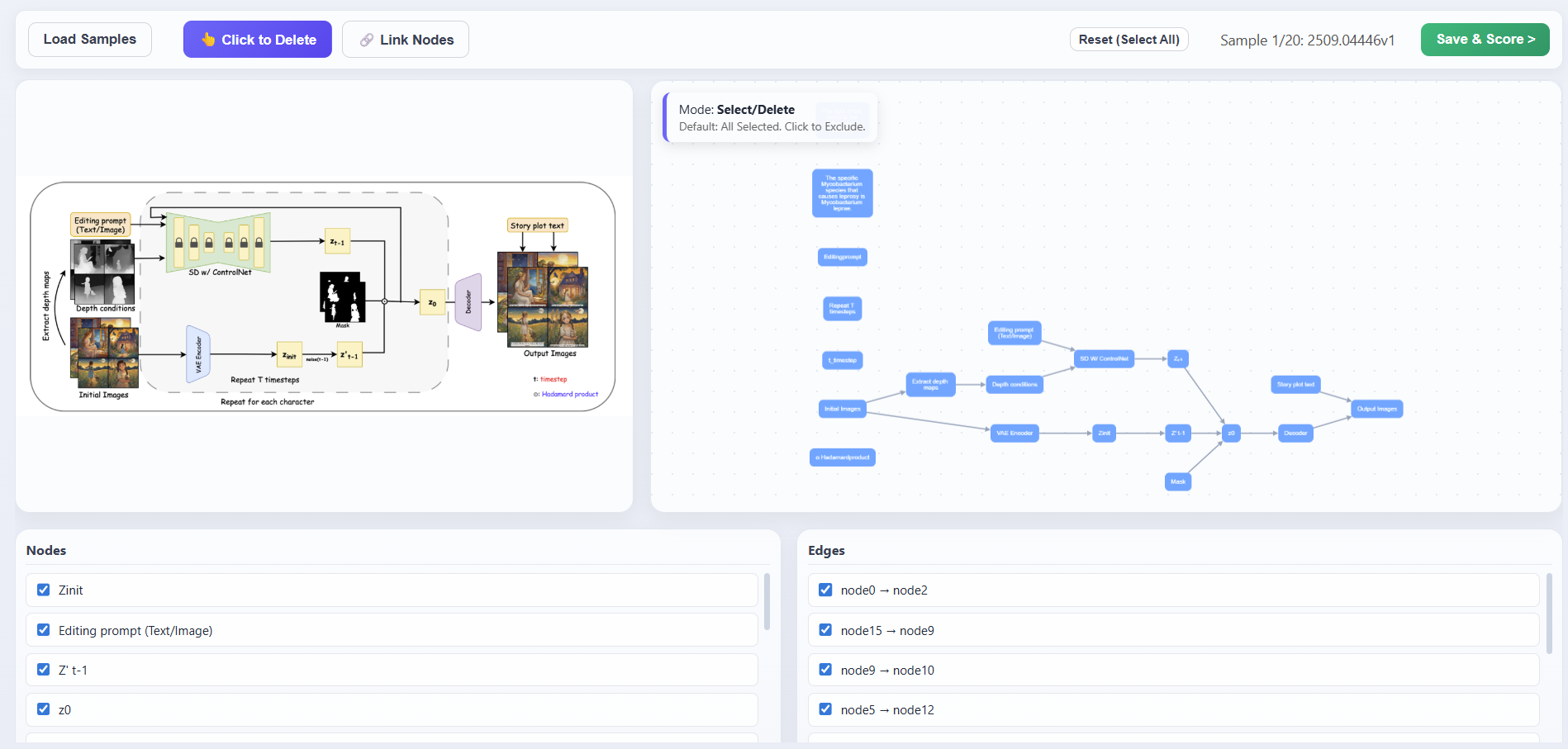}
    \vspace{-0.5em}
    {\small\textbf{(a)} Original graph automatically extracted by the HMAS pipeline before human verification.}
  \end{minipage}

  \vspace{1em}

  \begin{minipage}[t]{0.95\columnwidth}
    \centering
    \includegraphics[width=\linewidth]{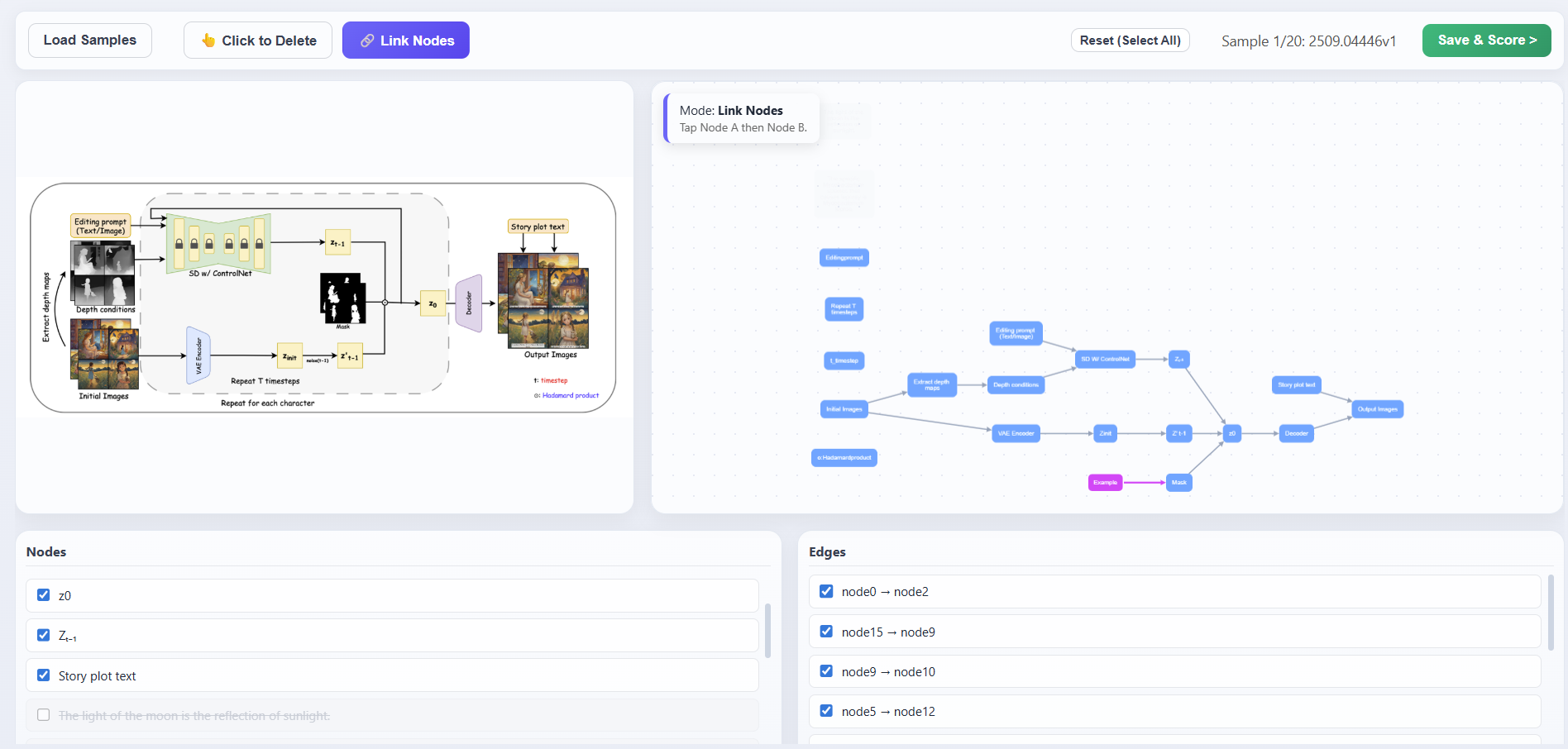}
    \vspace{-0.5em}
    {\small\textbf{(b)} Human-verified graph after minimal-intervention editing. Newly added elements are highlighted in magenta.}
  \end{minipage}

  \caption{\textbf{Human verification interface and representative annotation outcomes.} Annotators refine the automatically extracted graph by selectively excluding unsupported components and adding missing nodes or relations under a minimal-intervention, identity-consistent editing protocol.}
  \label{fig:appendix_ui}
\end{figure}

\subsection{Visual Encoding and Feedback}

To facilitate reliable and low-error annotation, the interface provides explicit visual encoding of element status.
Nodes and edges inherited from the automated extraction are displayed in blue.
Newly added nodes and edges introduced during verification are highlighted in magenta.
Excluded elements are visually muted and removed from active selection.

This color-coding provides immediate feedback on the scope and nature of human intervention.
Annotators can easily identify which parts of the graph were modified, reducing accidental edits and enabling quick review before final submission.

\subsection{Annotation Workflow and Quality Control}

Each source framework figure is verified independently by a trained annotator with a background in computer science.
Annotators are instructed to rely solely on visual evidence present in the diagram, without consulting the accompanying paper text beyond what is explicitly rendered in the figure.

To ensure consistency, annotators follow a shared annotation guideline that defines common error categories, including unsupported nodes, duplicated components, ambiguous relations, and missing dependencies.
Edge cases involving inherently ambiguous or implicit relations are resolved conservatively, favoring exclusion unless a dependency is clearly indicated.

All verification sessions are logged, including the number of removed and added elements, total annotation time, and edit history.
These logs enable post-hoc inspection and help ensure that annotation behavior remains consistent across domains and samples.

\subsection{Qualitative Examples}

Figure~\ref{fig:appendix_ui} presents representative examples of the human verification process.
Differences between the automatic and human-verified graphs primarily reflect the removal of weakly supported connections and the addition of missing but semantically necessary relations.
These discrepancies are most commonly observed in regions with dense layouts, overlapping visual elements, or implicit arrow semantics.

Together with the quantitative agreement results in Table~\ref{tab:human_verification}, these qualitative examples demonstrate that the automated HMAS pipeline recovers the majority of structural components correctly.
Remaining errors are sparse, interpretable, and amenable to targeted human correction, supporting the reliability of SciFlow-Bench as a benchmark grounded in realistic scientific diagrams.

\section{Discrepancy Analysis}
\label{sec:appendix_discrepancy}

This appendix provides a qualitative analysis of representative discrepancies observed during human verification of canonical ground-truth graphs.
The purpose of this analysis is not to enumerate failure cases exhaustively, but to clarify the dominant sources of mismatch between automatic extraction and human-refined graphs, and to contextualize these discrepancies within the inherent ambiguity

\begin{figure}[H]
\centering
\includegraphics[width=\columnwidth]{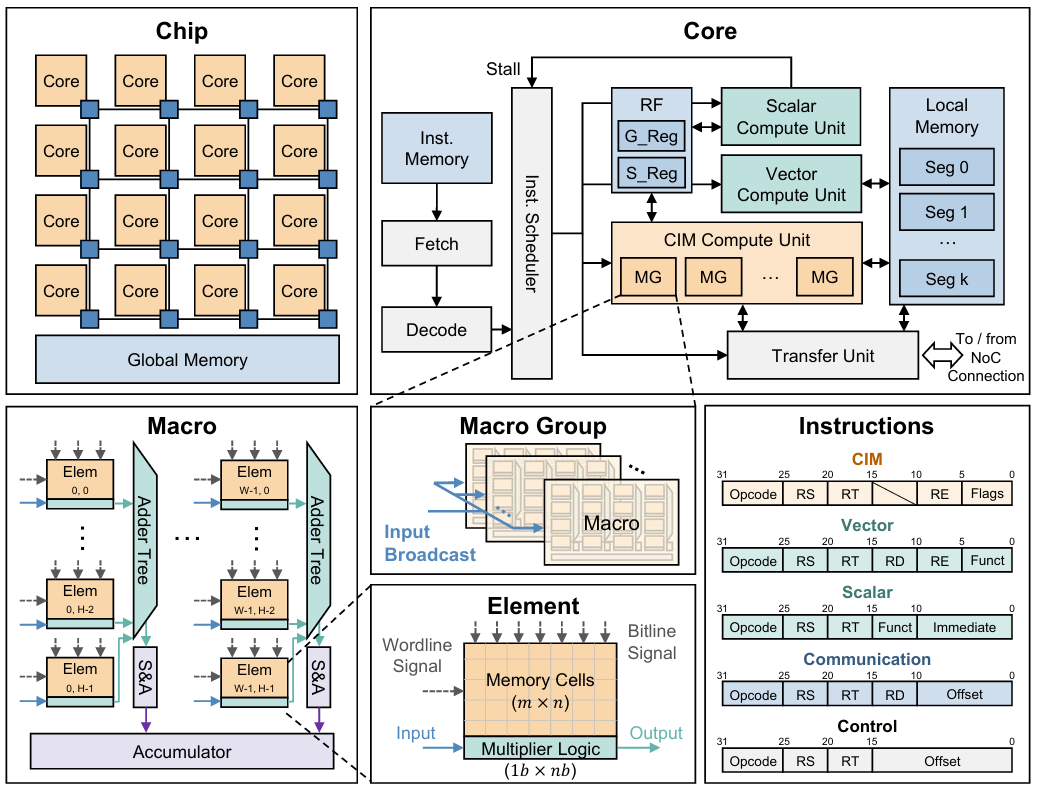}

\vspace{0.6em}

\includegraphics[width=\columnwidth]{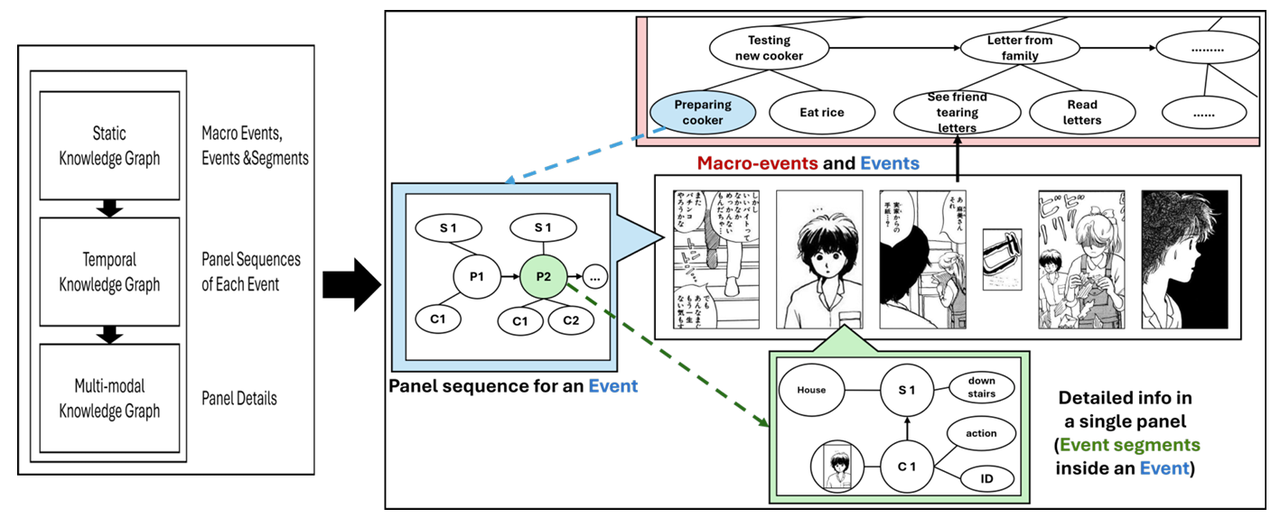}

\vspace{0.6em}

\includegraphics[width=\columnwidth]{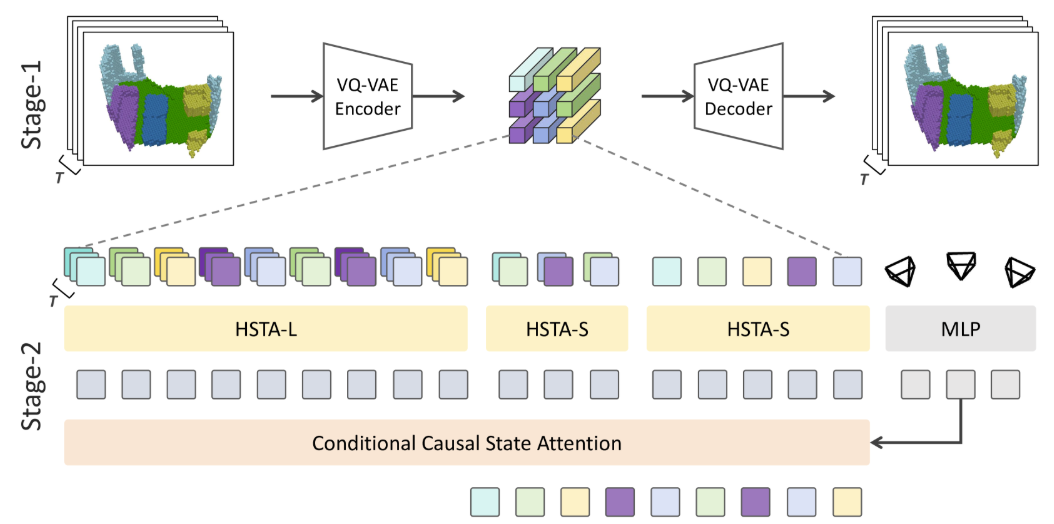}

\caption{\textbf{Representative discrepancy cases observed during human verification.} Examples illustrate common sources of mismatch between automatic extraction and human-verified ground-truth graphs, including ambiguous or implicit connections, text-related perception errors, and structurally ambiguous relations inherent in real scientific diagrams. Figures are reproduced from prior work~\cite{qi2025cimflow, chen2025structured, zhang2025occupancy}.}
\label{fig:error_cases}
\end{figure}

of real scientific diagrams.

Importantly, the observed mismatches do not indicate systematic failure of the HMAS pipeline.
Instead, they largely reflect intrinsic challenges in interpreting complex, densely packed, or weakly specified diagram structures that are also difficult for human readers without domain context.
We group the observed discrepancies into three recurring categories, illustrated in Figure~\ref{fig:error_cases}.

\subsection{Ambiguous or Implicit Relations}

A common source of discrepancy arises from relations that are visually suggested but not explicitly specified.
Examples include dashed arrows, unlabeled connectors, bidirectional links without clear causal direction, or graphical proximity implying association rather than dependency.
In such cases, human annotators may infer intent based on surrounding context or prior knowledge, while the automated pipeline conservatively avoids asserting unsupported directed edges.

This phenomenon is particularly prevalent in high-level architectural diagrams, where authors intentionally omit explicit semantics to improve visual clarity.
As shown in the top examples of Figure~\ref{fig:error_cases}, dashed or stylistic links often encode informal relationships that admit multiple plausible interpretations.
From an evaluation standpoint, preserving such ambiguity is preferable to hallucinating precise dependencies.
Accordingly, these discrepancies reflect a deliberate trade-off favoring structural faithfulness over speculative inference.

\subsection{Text-Related Perception Errors}

A second major class of discrepancies stems from text perception and grounding limitations.
These include vertically oriented labels, small subscripts, dense textual clusters, or stylized fonts that challenge optical character recognition and semantic association.
While humans can often resolve such cases by visual intuition or domain familiarity, automated systems rely on explicit textual signals that may be incomplete or noisy.

The middle examples in Figure~\ref{fig:error_cases} illustrate cases where missing or partially recognized text leads to node omission or incorrect semantic labeling, which in turn propagates to downstream edge mismatches.
These errors are well-documented in prior text-centric visual understanding work~\cite{chen2023textdiffuser} and remain an open challenge for multimodal perception systems.
Notably, such discrepancies tend to be sparse and localized rather than structural, indicating that the majority of diagram content is correctly grounded.

\subsection{Structurally Underspecified or Symbolic Nodes}

The third discrepancy category involves nodes or relations that are intentionally abstract or symbolic.
Common examples include identifier-based components such as \texttt{S1}, \texttt{P1}, or \texttt{C1}, purely graphical placeholders, or schematic elements whose semantics are defined only implicitly in the accompanying text.
Without explicit visual grounding, both humans and machines must rely on external context to interpret these elements.

As shown in the bottom examples of Figure~\ref{fig:error_cases}, annotators may choose to merge, reinterpret, or reattach such nodes during verification, while the automatic pipeline treats them conservatively as isolated or weakly connected entities.
These discrepancies highlight a fundamental limitation of diagram-only interpretation and underscore the necessity of structure-aware evaluation that tolerates multiple valid graph realizations.

\subsection{Summary and Implications}

Across all analyzed cases, discrepancies are sparse, interpretable, and systematically attributable to diagram ambiguity rather than parser instability.
Crucially, they do not exhibit cascading or catastrophic structural failure, and most corrections involve local edge refinement or limited node adjustment.
This observation aligns with the quantitative agreement results reported in Table~\ref{tab:human_verification}, where node-level recall remains high and edge-level errors concentrate in structurally ambiguous regions.

Taken together, this analysis supports the reliability of SciFlow-Bench as a benchmark grounded in realistic scientific diagrams.
The remaining mismatches reflect precisely the types of ambiguity that motivate structure-first evaluation, and further justify the need for inverse parsing protocols that explicitly reason about recoverable structure rather than surface-level visual fidelity alone.

\section{Qualitative Comparison of Generated Diagrams}
\label{sec:appendix_qualitative}

This appendix presents qualitative comparisons of scientific diagram images generated under identical structured visual prompts.
All examples are produced from the same method descriptions and evaluated under the same pixel-based generation and inverse parsing pipeline, without any manual prompt tuning or post-processing.

The purpose of this qualitative analysis is not to assess aesthetic quality or visual appeal in isolation.
Instead, these examples are provided to facilitate qualitative inspection of \emph{structural recoverability}, namely whether a generated diagram image preserves sufficient visual and semantic cues to support reliable reconstruction of an explicit graph representation under the inverse parsing protocol defined in SciFlow-Bench.

\begin{figure}[t]
\centering

\includegraphics[width=\columnwidth]{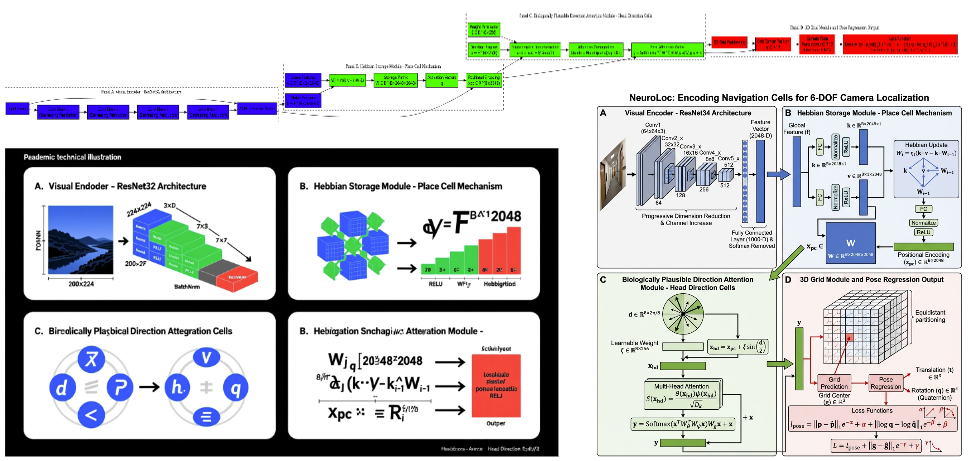}\\
{\small Case~1}\\[1.5ex]

\includegraphics[width=\columnwidth]{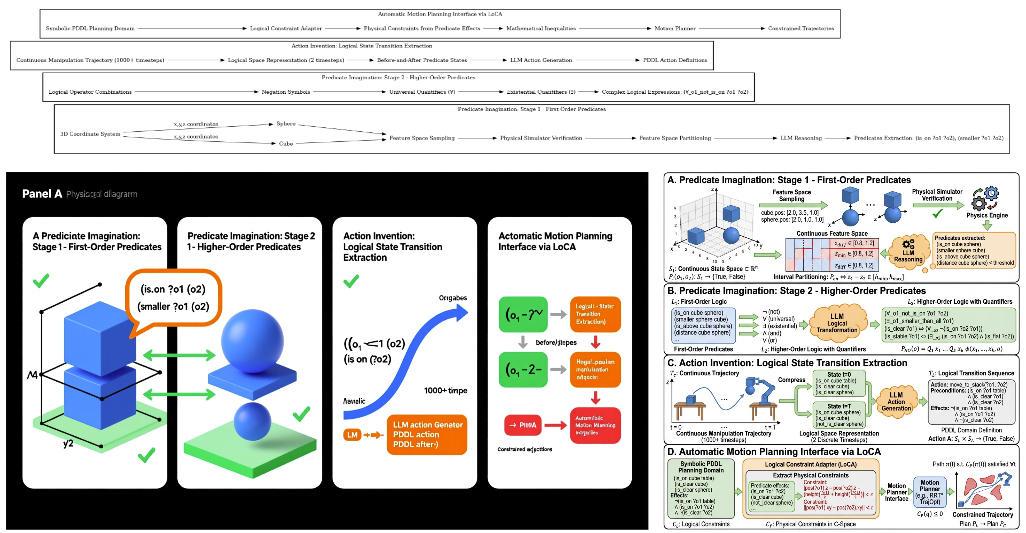}\\
{\small Case~2}

\caption{\textbf{Qualitative comparison of scientific diagram generation under identical structured visual prompts.}
For each case, the top diagram is a layout-driven Graphviz reference, while the bottom-left and bottom-right diagrams are generated by Qwen-Image and Gemini~3~Pro~Image, respectively.
All images are rendered at the same resolution and evaluated without manual tuning.
These examples are provided to illustrate differences in structural recoverability rather than aesthetic quality alone.}
\label{fig:appendix_qualitative}
\end{figure}

\subsection{Generation Paradigms}

We compare three representative generation paradigms that reflect fundamentally different assumptions about structure and rendering:

\begin{itemize}
  \item \textbf{Layout-driven reference (Graphviz).}
  A code-driven pipeline in which GPT-4o generates DOT code from the method description and Graphviz deterministically renders the diagram.
  This paradigm provides explicit topology and deterministic layout, and is included as a structural reference rather than a pixel-based competitor.

  \item \textbf{Diffusion-based generation (Qwen-Image).}
  An open-source diffusion model that operates purely in pixel space, representing the dominant paradigm for text-to-image generation without explicit structural supervision.

  \item \textbf{Autoregressive vision--language generation (Gemini~3~Pro~Image).}
  A commercial autoregressive vision--language model that jointly models text and vision during generation, representing the strongest pixel-based baseline evaluated in this work.
\end{itemize}

All models receive the same structured visual prompt produced by the cognitive planning layer, ensuring that observed differences arise from generation behavior rather than prompt variation.

\subsection{Qualitative Examples}

Figure~\ref{fig:appendix_qualitative} presents representative examples spanning different diagram styles and levels of structural complexity.
In each case, the Graphviz output provides an explicit and deterministic representation of functional modules and directed dependencies, serving as a diagnostic reference for topology.

The Qwen-Image outputs typically exhibit strong local visual coherence and stylistic consistency.
However, structural issues frequently arise, including missing intermediate components, ambiguous arrow terminations, or visually plausible but semantically unsupported connections.
These issues are often subtle at the pixel level yet lead to degraded graph-level recoverability under inverse parsing.

In contrast, the Gemini~3~Pro~Image outputs more often preserve interpretable layout cues, consistent arrow directionality, and stable module grouping.
As a result, they are more likely to be successfully inverse-parsed into structured graphs that align with the canonical ground-truth representation, particularly for diagrams with complex multi-stage pipelines.

\subsection{Structural Recoverability under Inverse Parsing}

A key observation from these examples is that visual plausibility alone does not guarantee structural recoverability.
Diffusion-based outputs may appear visually polished yet fail to encode sufficient relational cues for reliable reconstruction.
Common failure patterns include visually merged components, ambiguous arrow geometry, and decorative graphical elements that resemble functional nodes without clear semantic grounding.

Conversely, diagrams that preserve clear spatial separation, consistent arrow orientation, and stable text-to-region alignment—even if visually simpler—are more likely to yield high graph-level scores.
These qualitative patterns closely mirror the quantitative gap observed between image-level and graph-level metrics in Section~\ref{sec:benchmark_results}, reinforcing the claim that structure-aware evaluation is necessary to expose failure modes that are invisible to image-centric metrics.

\subsection{Discussion}

These qualitative comparisons complement the quantitative evaluation results by providing visual context under identical prompt conditions.
They highlight the fundamental trade-off between visual flexibility and structural determinism across generation paradigms.

Layout-driven methods guarantee explicit topology but sacrifice stylistic freedom.
Pixel-based models offer richer visual expressiveness but vary substantially in whether their outputs support reliable structural reconstruction.
Among pixel-based systems, stronger vision--language integration correlates with improved structural recoverability, although significant gaps remain.

Overall, these examples reinforce the central motivation of SciFlow-Bench:
scientific diagrams must be evaluated not only by how they look, but by whether their structure can be reliably recovered and interpreted.
Qualitative inspection under controlled prompt conditions provides critical insight into this distinction and supports the necessity of inverse parsing--based evaluation for scientific diagram generation.

\end{document}